\documentclass[10pt,twocolumn,letterpaper]{article}

\usepackage{cvpr}              

\usepackage{tikz-cd}
\usepackage[table]{xcolor}
\usepackage{array}
\usepackage{algorithm}
\usepackage{algpseudocode}

\definecolor{cvprblue}{rgb}{0.21,0.49,0.74}
\usepackage[pagebackref,breaklinks,colorlinks,allcolors=cvprblue]{hyperref}

\def\paperID{43} 
\def\confName{CVPR workshop}
\def\confYear{2026}

\title{SceneFunRI: Reasoning the Invisible for Task-Driven Functional Object Localization}

\author{
Posheng Chen$^{1}$, Powen Cheng$^{1}$, Gueter Josmy Faure$^{1}$, Hung-Ting Su$^{2}$, Winston H. Hsu$^1$\\
$^{1}$National Taiwan University\\
$^{2}$Delta Robotics Innovation Center \\
{\tt\small poshengchen@cmlab.csie.ntu.edu.tw}
}

\begin{document}
\maketitle
\begin{abstract} 
In real-world scenes, target objects may reside in regions that are not visible. While humans can often infer the locations of occluded objects from context and commonsense knowledge, this capability remains a major challenge for vision-language models (VLMs). To address this gap, we introduce \textbf{SceneFunRI}, a benchmark for \textbf{Reasoning the Invisible}. Based on the SceneFun3D dataset, SceneFunRI formulates the task as a 2D spatial reasoning problem via a semi-automatic pipeline and comprises 855 instances. It requires models to infer the locations of invisible functional objects from task instructions and commonsense reasoning. The strongest baseline model (Gemini 3 Flash) only achieves an CAcc@75 of 15.20, an mIoU of 0.74, and a Dist of 28.65. We group our prompting analysis into three categories: Strong Instruction Prompting, Reasoning-based Prompting, and Spatial Process of Elimination (SPoE). These findings indicate that invisible-region reasoning remains an unstable capability in current VLMs, motivating future work on models that more tightly integrate task intent, commonsense priors, spatial grounding, and uncertainty-aware search.
\end{abstract}

\section{Introduction}
\label{sec:intro}

\begin{figure*}[t]
  \centering
  \includegraphics[width=\textwidth]{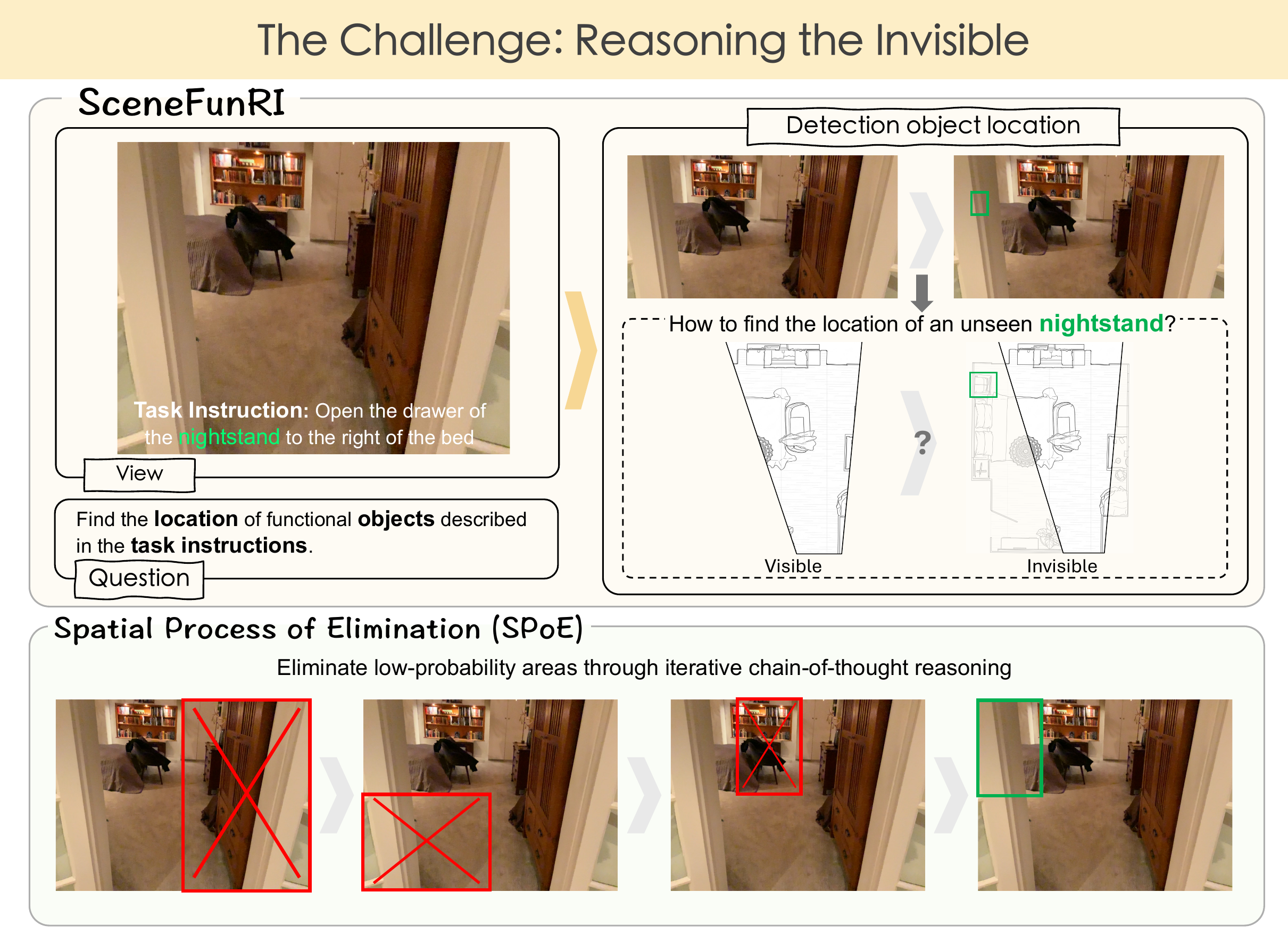}
  \caption{\textbf{Overview of SceneFunRI.} SceneFunRI introduces a benchmark for Reasoning the Invisible challenge, requiring models to localize task-relevant functional objects that are not directly visible. It evaluates whether models can infer hidden target locations from limited visual observations, task instructions, and commonsense reasoning. We further analyze Spatial Process of Elimination (SPoE), an iterative chain-of-thought framework that diagnoses model limitations by progressively eliminating unlikely regions.
}
  \label{fig:overview}
\end{figure*}

Spatial reasoning is a fundamental capability to understand and interact with the physical world.\cite{malanchini2020evidence}. This competence plays a central role in a wide range of applications, including robotics, autonomous driving, embodied agents, augmented and virtual reality \cite{song2025robospatial, tian2025nuscenes, yang2025embodiedbench}. In real-world environments, spatial reasoning is rarely conducted under ideal visual conditions. Objects are frequently partially occluded, truncated by image boundaries, or viewed from ambiguous perspectives. Therefore, occlusion is a major challenge for reliable perception and reasoning\cite{hsiao2014occlusion, saleh2021occlusion}. 

Although prior work has advanced spatial reasoning and occlusion understanding, existing benchmarks still fall short of real-world needs. Early spatial reasoning datasets mainly focus on explicit, localized, and often single-hop relations, such as object-to-object spatial recognition and controlled relational grounding \cite{SpatialSense, Rel3D, vsr23, kamath2023whatsup, COCO-Spatial, GQA-Spatial}, while even more comprehensive benchmarks that expand to navigation, perspective-taking, multi-step reasoning, and occlusion \cite{MindtheGap,omnispatial25,Capture,MultihopSpatial,spatialab} still tend to evaluate these abilities as largely disentangled sub-skills rather than within an integrated, task-driven setting. Meanwhile, research on occlusion has primarily centered on amodal completion \cite{ao2023image, ao2025open, xu2024amodal} and amodal segmentation \cite{ling2020variational, gao2023coarse, sun2022amodal, xiao2021amodal, jin2024llms, tai2025segment}, with recent efforts beginning to explore more challenging forms of reasoning over partially hidden content \cite{li2025aura,Capture}. However, these settings are still mostly guided by partially observable evidence, leaving fully occluded scenarios underexplored. As a result, existing benchmarks remain limited in assessing whether models can jointly integrate spatial understanding, commonsense knowledge, and contextual inference. This integration is necessary for reasoning about entirely invisible objects under incomplete observations, which is a core requirement for real-world decision-making.

This paper proposes SceneFunRI, a benchmark specifically designed for Reasoning the Invisible (RI). Built upon the SceneFun3D \cite{delitzas2024scenefun3d} dataset, SceneFunRI formulates RI as a two-dimensional spatial reasoning task through a semi-automatic procedure, resulting in 855 instances. RI introduces a new spatial reasoning challenge in which the target functional object is entirely invisible and provides no indirect visual cues. Unlike amodal or occlusion reasoning tasks, which rely on visible parts or indirect target-related cues, RI requires models to infer the target location solely from visible reference objects, task instructions, and commonsense knowledge. In real-world task execution, target objects are often not directly visible. Humans can construct a mental model \cite{johnson1980mental, johnson1983mental} of the surrounding environment from partial visual information and commonsense knowledge, thereby inferring the likely location of the target object. Although this task is intuitive for humans, current vision-language models (VLMs) still struggle with such implicit spatial reasoning.

To investigate this issue in greater depth, we group our prompting analysis into three categories. Strong Instruction Prompting constrains model output behavior through explicit and strict instructions, thereby reducing generative ambiguity. Reasoning-based Prompting (CoT and CoT-SR), in contrast, encourages the model to engage in step-by-step reasoning for more deliberate inference. Finally, we propose Spatial Process of Elimination (SPoE), which simulates human decision-making in spatial search tasks by progressively ruling out implausible candidate regions. Through this comparison, we aim to clarify the bottlenecks of VLM reasoning under the novel challenge of RI. 

The primary contributions of this work are summarized as follows:
\begin{itemize}
    \item We introduce a new challenge \textbf{RI} and propose \textbf{SceneFunRI} as a benchmark for evaluation, extending the focus of research from visible reasoning to invisible reasoning and thereby broadening the study of spatial reasoning.
    
    \item We propose \textbf{SPoE}, a localization analysis strategy that progressively eliminates less plausible regions.
    
    \item We analyze three different prompting strategies to better understand the reasoning bottlenecks of current VLMs in RI.

\end{itemize}

\section{Related Work}
\label{sec:related}

\begin{table*}[t] 
\centering
\renewcommand{\arraystretch}{1.2}
\setlength{\tabcolsep}{8pt}
\begin{tabular}{|p{0.14\textwidth}|p{0.34\textwidth}|p{0.41\textwidth}|}
\hline
\textbf{Occlusion Level} & \textbf{Definition} & \textbf{Example} \\
\hline
None & The target object is fully visible.
 & A car, person, or box is fully visible.
 \\
\hline
Partial Occlusion \cite{tai2025segment}
 & Part of the target object is visible, while the rest is occluded.
 & A car is partially blocked by another car.
 \\
\hline
Full Occlusion \cite{omnispatial25, Capture}
 & The target object is not directly visible, but \textcolor{red!70!black}{\textbf{indirect visual cues}} are present.
 & A person behind a wall is inferred from a shadow.
 \\
\hline
\textbf{Invisible (our)} 
 & The target object is not directly visible, and \textcolor{red!70!black}{\textbf{no indirect visual cues}} are available.
 & As shown in Figure \ref{fig:overview}, the target position is inferred from the instruction and surrounding objects.
 \\
\hline
\end{tabular}
\caption{\textbf{Occlusion Level.} We categorize occlusion levels into None, Partial Occlusion, Full Occlusion, and Invisible (ours). Among cases where the target object is completely not directly visible, the key distinction between Full Occlusion and Invisible lies in whether indirect visual cues are present.}
\label{tab:occlusion_level}
\end{table*}

\paragraph{Spatial Reasoning Benchmarks} 
Early benchmarks for spatial reasoning primarily focus on localized and explicit spatial relation recognition. For instance, SpatialSense \cite{SpatialSense} targets the identification of spatial relationships between objects in images, while Rel3D \cite{Rel3D} further incorporates 3D annotations to support spatial grounding. With the rise of vision-language models, VSR \cite{vsr23} introduces over 10k image-text pairs and 66 spatial relations to systematically evaluate models’ understanding of directionality, reference frames, and relational expressions. Similarly, datasets such as What’sUp \cite{kamath2023whatsup}, COCO-Spatial \cite{COCO-Spatial}, and GQA-Spatial \cite{GQA-Spatial} employ controlled settings to isolate fundamental spatial reasoning capabilities. While these benchmarks effectively reveal limitations in basic spatial judgments (e.g. left/right, front/behind, and orientation), they largely decompose spatial reasoning into single-hop, and explicitly expressed relations.

More recent efforts attempt to broaden the scope of spatial reasoning evaluation. On one hand, several works adopt a more comprehensive perspective on spatial cognition. Mind the Gap \cite{MindtheGap} unifies multiple dimensions including spatial relations, navigation, mental rotation, and spatial visualization. OmniSpatial \cite{omnispatial25} further extends this paradigm by incorporating dynamic reasoning, complex spatial logic, spatial interaction, and perspective-taking. On the other hand, emerging benchmarks begin to emphasize occlusion, compositionality, and multi-step reasoning. CAPTURE \cite{Capture} evaluates models’ ability to reason about visible and occluded objects through tasks such as counting and pattern recognition. MultihopSpatial \cite{MultihopSpatial} explicitly addresses the limitation of prior datasets by introducing multi-hop queries that require chaining spatial relations. SpatiaLab \cite{spatialab} evaluates spatial reasoning in realistic, open-world scenarios across diverse task categories, including depth and occlusion, spatial navigation, and 3D geometry, and introduces challenging subtasks such as Complete Occlusion Inference, which require reasoning about fully invisible objects.

Despite these advances, existing benchmarks still tend to treat spatial reasoning as a set of disentangled sub-skills, often separating it from other cognitive components. Even in more recent benchmarks, occlusion reasoning, navigation, and geometric understanding are typically evaluated in isolation, rather than within a combination of multiple abilities, task-driven framework. As a result, these benchmarks remain limited in their ability to reflect real-world scenarios, where reasoning often involves incomplete observations, implicit inference, and the integration of multiple cognitive abilities.


Prior complete-occlusion tasks often focus on cases where the target is not directly visible but can still be inferred from target-specific indirect cues, such as shadows, reflections, visible continuations, or occlusion contours. In contrast, SceneFunRI emphasizes a more challenging setting in which such target-specific cues are absent, and the target location must be inferred from scene-level contextual cues, including surrounding objects, scene layout, task instructions, and commonsense knowledge. This design makes SceneFunRI particularly suitable for evaluating functional object localization in real-world embodied task execution.

\paragraph{Occlusion Reasoning} 
Occlusion is a pervasive challenge in visual perception. Objects often obscure one another, making it difficult for models to understand a scene. To address this issue, previous research has explored amodal completion \cite{ao2023image, ao2025open, xu2024amodal} and amodal segmentation \cite{ling2020variational, gao2023coarse, sun2022amodal, xiao2021amodal, jin2024llms, tai2025segment}. The former focuses on inferring the complete shape of an object from partially visible information, while the latter aims to segment the full extent of objects in an image, including their invisible regions.

Recent studies have further extended this line of work to more challenging tasks. For example, AURA \cite{li2025aura} goes beyond amodal segmentation by incorporating reasoning about user interactions and implicit user intent. Meanwhile, CAPTURE \cite{Capture} introduces the problem of counting amodally, which involves counting objects arranged in patterns and requiring models to infer how these patterns continue behind occlusions.

However, existing amodal reasoning datasets and benchmarks primarily focus on scenarios involving partial occlusion, where reasoning is guided by partially observable visual evidence. In contrast, real-world environments frequently involve cases of full occlusion, where target objects are entirely invisible, making inference significantly more challenging. Current benchmarks lack a systematic evaluation of such reasoning capabilities under complete invisibility. To address this limitation, we propose a novel task termed RI, which aims to assess a model’s ability to infer the location of invisible objects.
\section{SceneFunRI Benchmark} 

\subsection{Overview} 


SceneFunRI consists of 855 task-driven localization instances derived from SceneFun3D. Each instance contains an RGB image, a task instruction, and a 2D ground-truth box corresponding to the projected location of an invisible functional object. The benchmark evaluates whether VLMs can infer the target location from visible contextual cues and commonsense knowledge.

\subsection{Reasoning the Invisible Definitions} 
The core objective of SceneFunRI is to evaluate the reasoning capability of VLMs when confronted with invisible objects. To this end, we impose the following task constraints: (1) the target object must be completely invisible in the image, (2) the target object must still lie within the image boundary, and (3) contextual cues must be retained. This design ensures that the model cannot rely on direct object detection. Instead, it must perform inference by leveraging task instructions, environmental context, spatial reasoning, and commonsense knowledge.

\subsection{Data Processing} 
\label{section:data_processing}
Our benchmark is built upon the \textbf{SceneFun3D} \cite{delitzas2024scenefun3d} dataset, utilizing its 3D point clouds, affordance annotations, task instructions, and scene video.  We processed 230 scenes from the training and validation splits, along with their corresponding videos, task descriptions, and ground-truth masks, and converted them into corresponding 2D spatial reasoning challenges:

\begin{enumerate} 
    \item \textbf{Frame Selection and Visibility Analysis:} For each scene, we generate a 3D mesh and compute object-to-image projections using two strategies: point-cloud projection based on camera intrinsics and extrinsics, and mesh-based projection that accounts for occlusion by scene geometry. From these projections, we derive two visibility measures: the projection visibility ratio ($V$) and the mesh-based visibility ratio ($V_o$). We retain frames with $V_o = 0\%$ and $V = 100\%$, indicating that the functional object exists in the scene coordinate system and lies within the camera view, but is fully occluded in the image.
    
    \item \textbf{Bounding Box Generation:} Ground-truth bounding boxes (BBoxes) are established by projecting the 3D point cloud annotations directly onto the 2D image plane based on the spatial extent of the hidden object.

\end{enumerate}
This automated pipeline initially yielded a candidate set of 1,232 data instances.

\subsection{Quality Control} 
\label{section:data_processing_result}
To improve the uniqueness, representativeness, and fairness of the benchmark, we conducted a second round of human quality control. We remove duplicate tasks, including repeated task instructions and duplicated images, as well as tasks with insufficient or ambiguous visual evidence. We retained a task only when the image contained sufficient contextual visual cues, the instruction did not correspond to multiple plausible regions in different directions, and the target location could be inferred from the image, task instruction, and commonsense knowledge. For example, if the instruction asks the agent to open the bedside table on the right side of the bed, the image should at least contain visual evidence indicating the presence and orientation of the bed. After filtering, the benchmark was reduced from 1,232 candidate tasks to 855 high-quality instances.
\section{Experiments Setup}
\subsection{Models} 
To comprehensively evaluate how the new tasks challenge model capabilities, we selected models that represent diverse architectural designs and training paradigms. These include \textbf{Gemini 3 Flash}\cite{team2023gemini}, \textbf{Qwen 3.5} series\cite{qwen3.5} (0.8B, 4B, 9B, 27B, and 122B), \textbf{InternVL 3.5} series\cite{wang2025internvl3_5} (1B, 4B, 8B, and 38B), \textbf{Gemma 3} series \cite{gemma_2025} (4B, 12B, and 27B), \textbf{Spatial Reasoning} RoboBrain2.5 8B \cite{tan2026robobrain}, 
MultiHopSpatial 4B \cite{MultihopSpatial}, and SenseNova-SI 8B \cite{sensenova-si}.

For reproducibility, all inference configurations were fixed for each model across experiments, including the checkpoint or API version, image resolution, prompt template, decoding parameters, maximum output length, and output parsing rules. Deterministic decoding was used whenever supported; otherwise, we fixed the random seed and all available generation parameters.

\subsection{Metrics} 


The evaluation metrics we adopt are CAcc@75, CAcc@50, mIoU, and Dist. Unlike standard object detection, Reasoning the Invisible does not require models to recover the exact visible contour of an object, since the target is entirely invisible and no object-specific visual evidence is available. In this setting, the predicted region should be interpreted as a search or interaction region rather than a tight visible-object bounding box. Therefore, IoU alone is overly strict: a prediction that safely contains the hidden object may receive a low IoU score if it is slightly larger than the projected target box. 

We therefore adopt a containment-based accuracy (CAcc@ $\theta$) criterion, which measures whether the hidden target lies within the predicted region. The bounding box should be as compact as possible to ensure precise localization. Consequently, we define the evaluation metric as follows:
\begin{equation*}
\textbf{CAcc@ $\theta$} :=
\left\{
\begin{array}{@{}l@{\quad}l@{}}
1 & \text{, if } b_{gt} \subset b_{p} \text{ and } 1 - \frac{A_p}{A_{img}} > \theta, \\
0 & \text{, otherwise.} \\
\end{array}
\right.
\end{equation*}
where $b_{p}$, $b_{gt}$ are the predicted and ground-truth regions. 
$A_p$, $A_{img}$ are the predicted regional area and the image area.
$\theta$ controls the compactness constraint on the predicted region. For example, CAcc@75 requires the prediction to contain the ground-truth box while occupying less than 25\% of the image area.

We use containment rather than exact box matching because the target object is invisible and its precise visible boundary cannot be directly observed by the model. Therefore, a prediction is considered successful if it covers the projected ground-truth extent of the hidden object while remaining sufficiently compact.

We additionally report Dist, a normalized corner-distance metric. Because IoU becomes zero even when the predicted box is spatially close to the ground-truth projection but does not overlap. Dist complements CAcc and mIoU by measuring localization proximity even in non-overlapping cases. The formula is defined as follows:
\begin{align*}
    \textbf{Dist} &:= \frac{1}{4}\sum_{i=1}^{2}\sum_{j=1}^{2}
\sqrt{
\left(\frac{x_i^p-x_i^g}{w}\right)^2+
\left(\frac{y_j^p-y_j^g}{h}\right)^2
},
\end{align*}
where the predicted bounding box is $[x_{1}^p, y_{1}^p, x_{2}^p, y_{2}^p ]$, the ground-truth bounding box is $[x_{1}^g, y_{1}^g, x_{2}^g, y_{2}^g ]$, and $h$ and $w$ denote the image height and width, respectively.

Note that all reported metric values are scaled by a factor of 100 for readability. CAcc@$\theta$ and mIoU are therefore reported in percentage points, while Dist denotes the normalized corner distance defined above multiplied by 100. For example, a reported Dist of 28.65 corresponds to an unscaled normalized distance of 0.2865.

\section{Results and Analysis}

\subsection{Overview Results}

\begin{table}[htbp] 
\centering
\renewcommand{\arraystretch}{1.1}
\setlength{\tabcolsep}{4pt}
\scriptsize
\resizebox{\linewidth}{!}{
\begin{tabular}{>{\raggedright\arraybackslash}p{2.3cm} r r r r}
\rowcolor[HTML]{F2F2F2}
\textbf{Model} & \textbf{CAcc@75}$\uparrow$ & \textbf{CAcc@50}$\uparrow$ & \textbf{mIoU}$\uparrow$ & \textbf{Dist}$\downarrow$ \\
\hline
\rowcolor[HTML]{CFE3EF}
\textbf{\textit{Qwen 3.5 Series}} & & & & \\
Qwen3.5-0.8B  & 3.63 & 5.73 & 0.06 & 79.70 \\
Qwen3.5-4B  & 10.99 & 14.27 & 0.31 & 38.93 \\
Qwen3.5-9B  & 11.11 & 12.87 & 0.31 & 45.03 \\
Qwen3.5-27B  & 7.95 & 10.29 & 0.34 & \textbf{\underline{38.90}} \\
Qwen3.5-122B  & \textbf{\underline{12.63}} & \textbf{\underline{16.14}} & \textbf{\underline{0.41}} & 39.76 \\
\rowcolor[HTML]{CFE3EF}
\textbf{\textit{InternVL3 Series}} & & & & \\
InternVL3-1B  & 3.27 & \textbf{\underline{4.44}} & 0.11 & 68.27 \\
InternVL3-4B  & 3.98 & 4.21 & 0.17 & 41.80 \\
InternVL3-8B  & \textbf{\underline{4.09}} & 4.21 & 0.25 & 45.66 \\
InternVL3-38B  & 3.86 & 4.21 & \textbf{\underline{0.34}} & \textbf{\underline{35.55}}\\
\rowcolor[HTML]{CFE3EF}
\textbf{\textit{Gemma 3 Series}} & & & & \\
gemma-3-4b-it  & 4.21 & 4.21 & 0.12 & 43.02 \\
gemma-3-12b-it  & 4.09 & 4.44 & 0.07 & \textbf{\underline{41.67}} \\
gemma-3-27b-it  & \textbf{\underline{7.60}} & \textbf{\underline{7.72}} & \textbf{\underline{0.12}} & 43.35 \\
\rowcolor[HTML]{CFE3EF}
\textbf{\textit{Spatial Reasoning}} & & & & \\
MultiHopSpatial 4B & \textbf{\underline{11.11}} & \textbf{\underline{13.33}} & 0.18 & 60.00 \\
SenseNova-SI 8B & 9.59 & 11.23 & 0.26 & \textbf{\underline{43.25}} \\
RoboBrain2.5 8B & 6.32 & 6.55 & \textbf{\underline{0.31}} & 44.52 \\
\rowcolor[HTML]{CFE3EF}
\textbf{\textit{Gemini Series}} & & & & \\
Gemini 3 Flash & \cellcolor[HTML]{FFF2CC}\textbf{\underline{15.20}} & \cellcolor[HTML]{FFF2CC}\textbf{\underline{17.19}} & \cellcolor[HTML]{FFF2CC}\textbf{\underline{0.74}} & \cellcolor[HTML]{FFF2CC}\textbf{\underline{28.65}} \\
\hline \hline
Human baseline & 66.43 & 66.43 & 3.40 & 13.39
\end{tabular}
}
\caption{\textbf{Overview of SceneFunRI Evaluation Results}. All values are multiplied by 100 for readability. CAcc@75, CAcc@50, and mIoU are reported in percentage points, while Dist is the normalized corner distance multiplied by 100. \textbf{\underline{Bold and underlined values}} denote the best in each series; \colorbox[HTML]{FFF2CC}{yellow highlights} mark the best overall.}
\label{table:base_prompt}
\end{table}

\begin{table*}[!t] 
\centering
\renewcommand{\arraystretch}{1.1}
\setlength{\tabcolsep}{4pt}
\scriptsize
\resizebox{\linewidth}{!}{
\begin{tabular}{>{\raggedright\arraybackslash}p{1.8cm} l r r r r | l r r r r}
\rowcolor[HTML]{F2F2F2}
\textbf{Model} & \textbf{Prompt} & \textbf{CAcc@75}$\uparrow$ & \textbf{CAcc@50}$\uparrow$ & \textbf{mIoU}$\uparrow$ & \textbf{Dist}$\downarrow$ & \textbf{Prompt} & \textbf{CAcc@75}$\uparrow$ & \textbf{CAcc@50}$\uparrow$ & \textbf{mIoU}$\uparrow$ & \textbf{Dist}$\downarrow$ \\
\hline
Qwen3.5-27B & Hidden & 16.73 \textcolor{green!50!black}{+8.78} & 21.40 \textcolor{green!50!black}{+11.11} & 0.33 \textcolor{red!70!black}{-0.00} & 37.08 \textcolor{green!50!black}{-1.82} & CoT & 16.84 \textcolor{green!50!black}{+8.89} & 19.88 \textcolor{green!50!black}{+9.59} & 0.42 \textcolor{green!50!black}{+0.08} & 34.34 \textcolor{green!50!black}{-4.56} \\
 & Common & 8.42 \textcolor{green!50!black}{+0.47} & 9.71 \textcolor{red!70!black}{-0.58} & 0.37 \textcolor{green!50!black}{+0.03} & 35.23 \textcolor{green!50!black}{-3.67} & CoT-SR & 14.04 \textcolor{green!50!black}{+6.09} & 16.73 \textcolor{green!50!black}{+6.44} & 0.38 \textcolor{green!50!black}{+0.04} & 37.18 \textcolor{green!50!black}{-1.72} \\
 & Both & 16.49 \textcolor{green!50!black}{+8.54} & 18.36 \textcolor{green!50!black}{+8.07} & 0.37 \textcolor{green!50!black}{+0.03} & 34.57 \textcolor{green!50!black}{-4.33} &  &  &  &  &  \\
\hline
gemma-3-27b-it & Hidden & 5.73 \textcolor{red!70!black}{-1.87} & 6.20 \textcolor{red!70!black}{-1.52} & 0.07 \textcolor{red!70!black}{-0.04} & 45.74 \textcolor{red!70!black}{+2.39} & CoT & 7.13 \textcolor{red!70!black}{-0.47} & 7.37 \textcolor{red!70!black}{-0.35} & 0.15 \textcolor{green!50!black}{+0.03} & 39.63 \textcolor{green!50!black}{-3.72} \\
 & Common & 7.02 \textcolor{red!70!black}{-0.58} & 7.49 \textcolor{red!70!black}{-0.23} & 0.15 \textcolor{green!50!black}{+0.03} & 41.88 \textcolor{green!50!black}{-1.47} & CoT-SR & 6.78 \textcolor{red!70!black}{-0.82} & 7.02 \textcolor{red!70!black}{-0.70} & 0.16 \textcolor{green!50!black}{+0.04} & 39.81 \textcolor{green!50!black}{-3.54} \\
 & Both & 4.91 \textcolor{red!70!black}{-2.69} & 5.38 \textcolor{red!70!black}{-2.34} & 0.09 \textcolor{red!70!black}{-0.02} & 47.98 \textcolor{red!70!black}{+4.63} &  &  &  &  &  \\
\hline
InternVL3\_5-38B & Hidden & 4.09 \textcolor{green!50!black}{+0.23} & 4.44 \textcolor{green!50!black}{+0.23} & 0.21 \textcolor{red!70!black}{-0.12} & 36.71 \textcolor{red!70!black}{+1.16} & CoT & 4.09 \textcolor{green!50!black}{+0.23} & 4.44 \textcolor{green!50!black}{+0.23} & 0.22 \textcolor{red!70!black}{-0.11} & 42.67 \textcolor{red!70!black}{+7.12} \\
 & Common & 3.51 \textcolor{red!70!black}{-0.35} & 3.74 \textcolor{red!70!black}{-0.47} & 0.16 \textcolor{red!70!black}{-0.17} & 36.21 \textcolor{red!70!black}{+0.66} & CoT-SR & 5.38 \textcolor{green!50!black}{+1.52} & 5.50 \textcolor{green!50!black}{+1.29} & 0.38 \textcolor{green!50!black}{+0.04} & 43.39 \textcolor{red!70!black}{+7.84} \\
 & Both & 3.16 \textcolor{red!70!black}{-0.70} & 3.16 \textcolor{red!70!black}{-1.05} & 0.23 \textcolor{red!70!black}{-0.10} & 36.52 \textcolor{red!70!black}{+0.97} &  &  &  &  &  \\
\hline
Qwen3.5-122B & Hidden & 15.32 \textcolor{green!50!black}{+2.69} & 19.06 \textcolor{green!50!black}{+2.92} & 0.28 \textcolor{red!70!black}{-0.12} & 41.89 \textcolor{red!70!black}{+2.13} & CoT & 15.79 \textcolor{green!50!black}{+3.16} & 20.23 \textcolor{green!50!black}{+4.09} & 0.37 \textcolor{red!70!black}{-0.03} & 37.41 \textcolor{green!50!black}{-2.35} \\
 & Common & 13.22 \textcolor{green!50!black}{+0.59} & 15.79 \textcolor{red!70!black}{-0.35} & 0.40 \textcolor{red!70!black}{-0.00} & 38.68 \textcolor{green!50!black}{-1.08} & CoT-SR & 16.02 \textcolor{green!50!black}{+3.39} & 19.30 \textcolor{green!50!black}{+3.16} & 0.38 \textcolor{red!70!black}{-0.02} & 35.19 \textcolor{green!50!black}{-4.57} \\
 & Both & 15.44 \textcolor{green!50!black}{+2.81} & 19.06 \textcolor{green!50!black}{+2.92} & 0.31 \textcolor{red!70!black}{-0.09} & 38.64 \textcolor{green!50!black}{-1.12} &  &  &  &  &  \\
\hline
Gemini 3 Flash & Hidden & 25.03 \textcolor{green!50!black}{+9.83} & 31.23 \textcolor{green!50!black}{+14.04} & 0.69 \textcolor{red!70!black}{-0.04} & 31.91 \textcolor{red!70!black}{+3.26} & CoT & 13.92 \textcolor{red!70!black}{-1.28} & 17.08 \textcolor{red!70!black}{-0.11} & 0.68 \textcolor{red!70!black}{-0.05} & 28.79 \textcolor{red!70!black}{+0.14} \\
 & Common & 5.26 \textcolor{red!70!black}{-9.94} & 5.61 \textcolor{red!70!black}{-11.58} & 0.74 \textcolor{green!50!black}{+0.00} & 28.06 \textcolor{green!50!black}{-0.59} & CoT-SR & 14.27 \textcolor{red!70!black}{-0.93} & 17.31 \textcolor{green!50!black}{+0.12} & 0.70 \textcolor{red!70!black}{-0.03} & 28.55 \textcolor{green!50!black}{-0.10} \\
 & Both & 19.77 \textcolor{green!50!black}{+4.57} & 22.34 \textcolor{green!50!black}{+5.15} & 0.72 \textcolor{red!70!black}{-0.01} & 28.64 \textcolor{green!50!black}{-0.01} &  &  &  &  &  \\
\end{tabular}
}
\caption{\textbf{Different Prompting Settings.} All metric values are scaled by 100. Left: Strong Instruction Prompting. Right: Reasoning-based Prompting. Green values indicate \textcolor{green!50!black}{improvements} over the baseline, while red values indicate performance \textcolor{red!70!black}{degradation} relative to the baseline.}
\label{table:sip_rb}
\end{table*}

Table \ref{table:base_prompt} presents the overall performance differences among various model families on the benchmark. The results can be further analyzed using metrics such as CAcc@75, CAcc@50, mIoU, and Dist, which collectively reflect differences in localization capability and strategic behavior. Overall, Gemini 3 Flash achieves the best model performance, attaining the highest CAcc@75 (15.20), CAcc@50 (17.19), and mIoU (0.74), while also maintaining the lowest Dist (28.65). By contrast, the human baseline achieves substantially higher CAcc@75 (66.43), CAcc@50 (66.43), and mIoU (3.40), as well as a lower Dist (13.39), indicating substantial room for model improvement. The low human mIoU despite high CAcc further supports our metric design: even humans tend to identify plausible search regions rather than precisely matching the projected hidden-object box.

Among open-source models, the Qwen 3.5 series generally performs better than other open-source model families. In particular, Qwen3.5-122B achieves the highest CAcc@75 (12.63), CAcc@50 (16.14), and mIoU (0.41), along with a relatively low Dist (39.76). Although its CAcc@50 approaches that of Gemini 3 Flash, noticeable gaps remain in CAcc@75, mIoU, and Dist. This suggests that the model tends to select larger regions, resulting in reduced precision.

In contrast, the InternVL3.5 series exhibits relatively low CAcc@75 and CAcc@50 scores, both approximately 4, indicating that only a limited number of tasks are successfully localized, with model size having little impact on overall performance. However, it performs comparatively well in terms of mIoU (0.34) and Dist (35.55). This implies that the model adopts a more conservative prediction strategy, achieving higher precision but lower accuracy.

Despite being fine-tuned for spatial reasoning tasks, specialized models such as MultiHopSpatial, SenseNova-SI, and RoboBrain2.5 do not outperform Qwen3.5-4B and Qwen3.5-9B across multiple metrics. This suggests that the proposed benchmark captures a challenging aspect of spatial reasoning that warrants further investigation.

\subsection{Prompting Analysis}
To analyze the challenges that this new problem and benchmark pose to existing models, we design three types of experiments for a comprehensive evaluation. Specifically, we investigate the effects of different prompting strategies: Strong Instruction Prompting, Reasoning-based Prompting, and Spatial Process of Elimination (SPoE).

\subsubsection{Strong Instruction Prompting} 
Since current models primarily rely on visible visual reasoning and are not trained to locate invisible objects through commonsense reasoning, we introduce additional instructions into the basic prompt. These instructions guide the model to reason about invisible regions and apply commonsense knowledge. Our prompting strategies are divided into the following types:

\begin{itemize}
    \item \textbf{Hidden:} The model is instructed to ignore all visible objects and perform reasoning only over invisible regions.
    
    \item \textbf{Common:} Many assumptions in daily life are considered common sense by humans but may not be understood by models. For example, ''Unplug outlets are often located in corners or behind furniture.'' Therefore, we explicitly include relevant commonsense knowledge in the prompt and require the model to reason based on it.
    
    \item \textbf{Both:} A combination of both the Hidden Prompt and Common Prompt added to the basic prompt.
\end{itemize}

When comparing the three prompting strategies (Hidden, Common, and Both) shown in Table \ref{table:sip_rb} left, clear differences in overall trends can be observed. The Hidden prompt consistently yields stable performance improvements across most models, with particularly notable gains in the Gemini and Qwen series. This suggests that models may possess latent reasoning capabilities that are not actively engaged when reasoning about invisible information.

In contrast, although the Common prompt provides prior knowledge about the locations of target objects, its effectiveness is limited. It even results in performance degradation on certain metrics. We hypothesize that this is due to a notable gap in current models between language understanding and spatial grounding. Even when the textual description is accurate, models often struggle to precisely translate semantic information into correct visual bounding box predictions, leading to localization errors that negatively impact overall performance.

As for the Both prompt (which combines Hidden and Common), it does not yield substantial improvement. The underlying reason appears similar to that of the Common prompt: although the language-based reasoning is correct, the mismatch between textual descriptions and spatial localization introduces errors, ultimately reducing overall performance.

\subsubsection{Reasoning-based Prompting} 
To better understand the role of prompting, we encourage the model to perform multi-step reasoning before producing the final answer. We design a three-step reasoning process:

\begin{enumerate}
    \item Analyze the functional object required for interaction based on the task instruction.
    \item Determine whether the target object is visible in the image.
    \item Locate the target object based on commonsense knowledge and visual cues.
\end{enumerate}

In addition to the standard CoT method, we further explore CoT with Self-Reflection (CoT-SR), where the model performs self-reflection based on the initial CoT results.

Table \ref{table:sip_rb} right presents the impact of reasoning-based prompting on model performance. Performance improvements are observed exclusively in the Qwen 3.5 27B and 122B models, while other models generally maintain their baseline performance.  We hypothesize that this discrepancy primarily stems from differences in automatic structured reasoning capabilities. For instance, models such as Gemini 3 inherently generate structured CoT reasoning, expanding intermediate reasoning steps during inference. As a result, the addition of CoT prompting does not yield further performance gains. In contrast, Qwen 3.5 models, when not operating in Thinking mode, do not automatically produce structured reasoning steps and therefore benefit from both CoT and CoT-SR prompting.

\subsubsection{Spatial Process of Elimination (SPoE)} 
To address the challenges posed by SceneFunRI, we propose the Spatial Process of Elimination (SPoE) to diagnose and clarify the limitations of models on this task. This approach is inspired by human reasoning and integrates elimination strategies, and Chain-of-Thought (CoT) reasoning.

Traditionally, the process of elimination is mainly applied to multiple-choice questions \cite{poe, balepur-etal-2024-easy}. However, in spatial reasoning tasks, the target object may not be within the current field of view. In such cases, its location cannot be directly inferred. Humans therefore tend to rule out less likely regions. This strategy improves both the efficiency and the likelihood of successfully locating the target.

\begin{algorithm}[htbp]
\caption{\textbf{Pseudocode of SPoE.}}
\label{alg:spoe}
\begin{algorithmic}[1]
\Require Image $I$, query $q$, model $M$, iterations $T$
\Ensure Localized target region $B^*$
\State $B \gets$ full image bounding box
\For{$t = 1$ to $T$}
    \State $\{B_1, B_2\} \gets \mathrm{Split}(B,\, t \bmod 2)$
    \State $B_{elim} \gets M(\mathrm{OverlayBBoxes}(I,\{B_1,B_2\}), q)$
    \State $\text{model selects the less likely region to eliminate}$
    \State $B \gets B \setminus B_{elim}$
\EndFor
\State \Return $B$
\end{algorithmic}
\end{algorithm}

Based on this intuition, SPoE progressively narrows the search space through iterative region elimination. We leverage the model’s ability to identify objects within bounding boxes (bboxes). Specifically, the target image is partitioned into two regions, and the corresponding bounding boxes are overlaid to prompt the model to choose between them, thereby eliminating the region with lower likelihood. Subsequently, the remaining candidate region is further divided into two subregions and this process is repeated. Through this recursive partitioning-and-elimination procedure, the search space is progressively reduced. Consequently, the model can incrementally localize the position of the hidden target.

In Algorithm~\ref{alg:spoe}, $\mathrm{Split}(B, t \bmod 2)$ denotes vertical splitting when $t$ is odd and horizontal splitting otherwise. $\mathrm{OverlayBBoxes}(I,\{B_1,B_2\})$ denotes overlaying the candidate bounding boxes $B_1$ and $B_2$ on image $I$ for model selection.

\begin{figure}[htbp]
    \centering
    \includegraphics[width=1\linewidth]{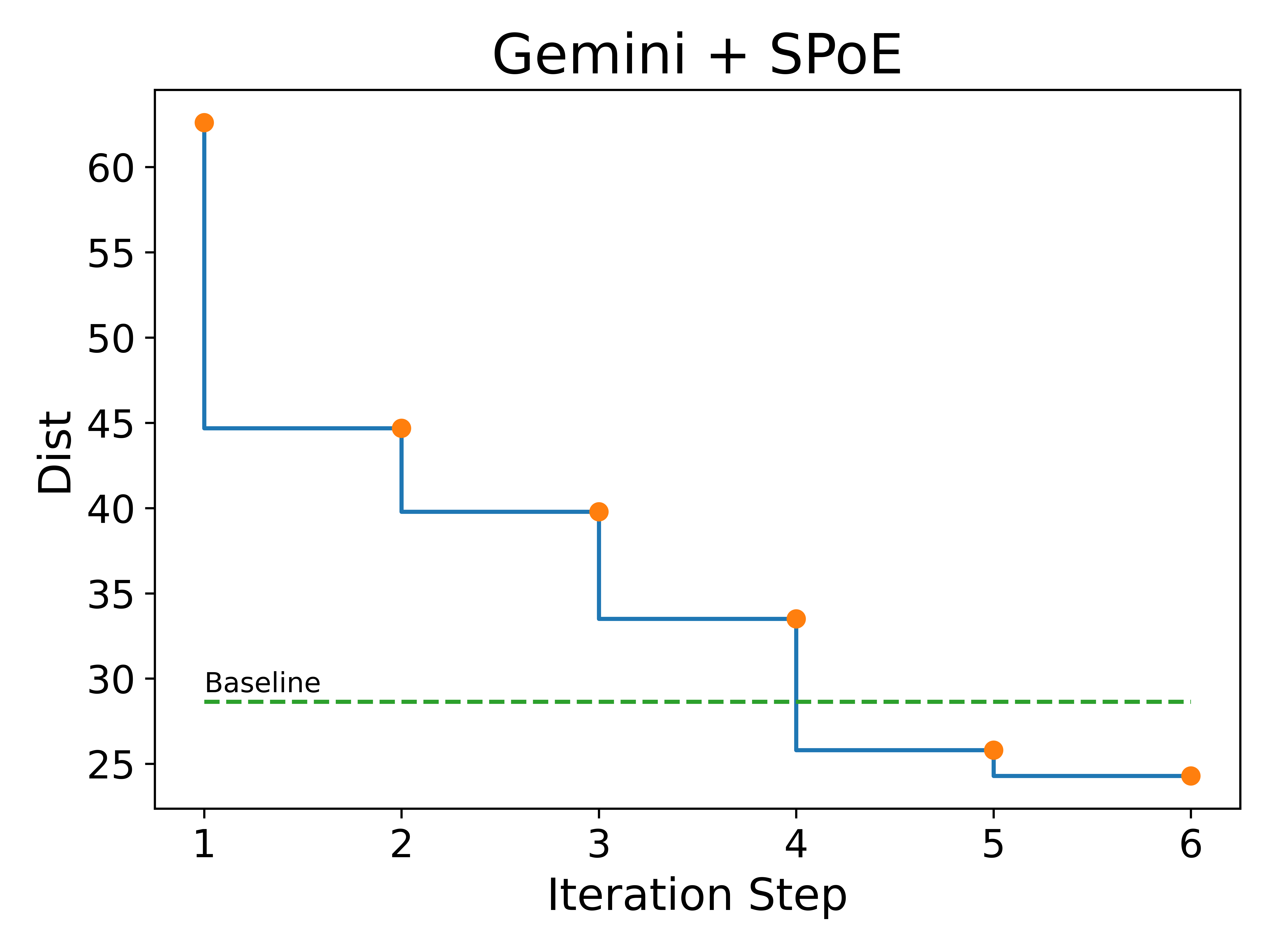}
    \caption{\textbf{SPoE analysis results.} As the number of iterations increases, the Dist metric decreases progressively, indicating that eliminating low-probability visible regions effectively narrows the search space. x-axis represents the iteration steps, and y-axis denotes the Dist value. The dashed line corresponds to the baseline score reported in Table \ref{table:base_prompt}, serving as a reference for comparison.}
    \label{fig:spoe}
\end{figure}

Figure \ref{fig:spoe} illustrates the effect of SPoE on the performance of Gemini 3 Flash. In this analysis, we mainly focus on the Dist metric, since CAcc@75, CAcc@50, and mIoU are all overlap-based metrics. For an iterative elimination strategy such as SPoE, the predicted region is progressively narrowed by removing candidate areas rather than directly optimizing the overlap between a predicted bounding region and the ground truth. As a result, these overlap-based metrics are less sensitive to the intermediate changes produced during the elimination process and may fail to reflect meaningful progress across iterations. In contrast, Dist directly measures the distance between the predicted location and the target, making it more suitable for observing whether the model gradually moves closer to the correct region.

The results show that the Dist value consistently decreases as the number of iterations increases, indicating that the model can leverage visible regions for iterative elimination. This process progressively narrows the candidate region and moves the predicted location closer to the target. Notably, after the fourth iteration, the Dist value falls below the baseline reported in Table \ref{table:base_prompt}, suggesting that, once the search space is sufficiently constrained, the model demonstrates a more explicit spatial localization capability. However, further observation reveals that, after a certain number of iterations, low-probability candidate regions are substantially reduced or even eliminated. Despite this, the model remains unable to determine an appropriate stopping point on its own and continues to exclude regions. This suggests that VLMs still have limitations in reasoning about uncertainty in invisible regions.

We argue that directly inferring hidden regions is a substantial challenge for models. Instead, a more effective approach may involve analyzing the observable regions and reasoning by excluding candidate areas with low probability.

\subsection{Qualitative results}

\begin{figure}[t]
  \centering
  \includegraphics[width=0.45\textwidth]{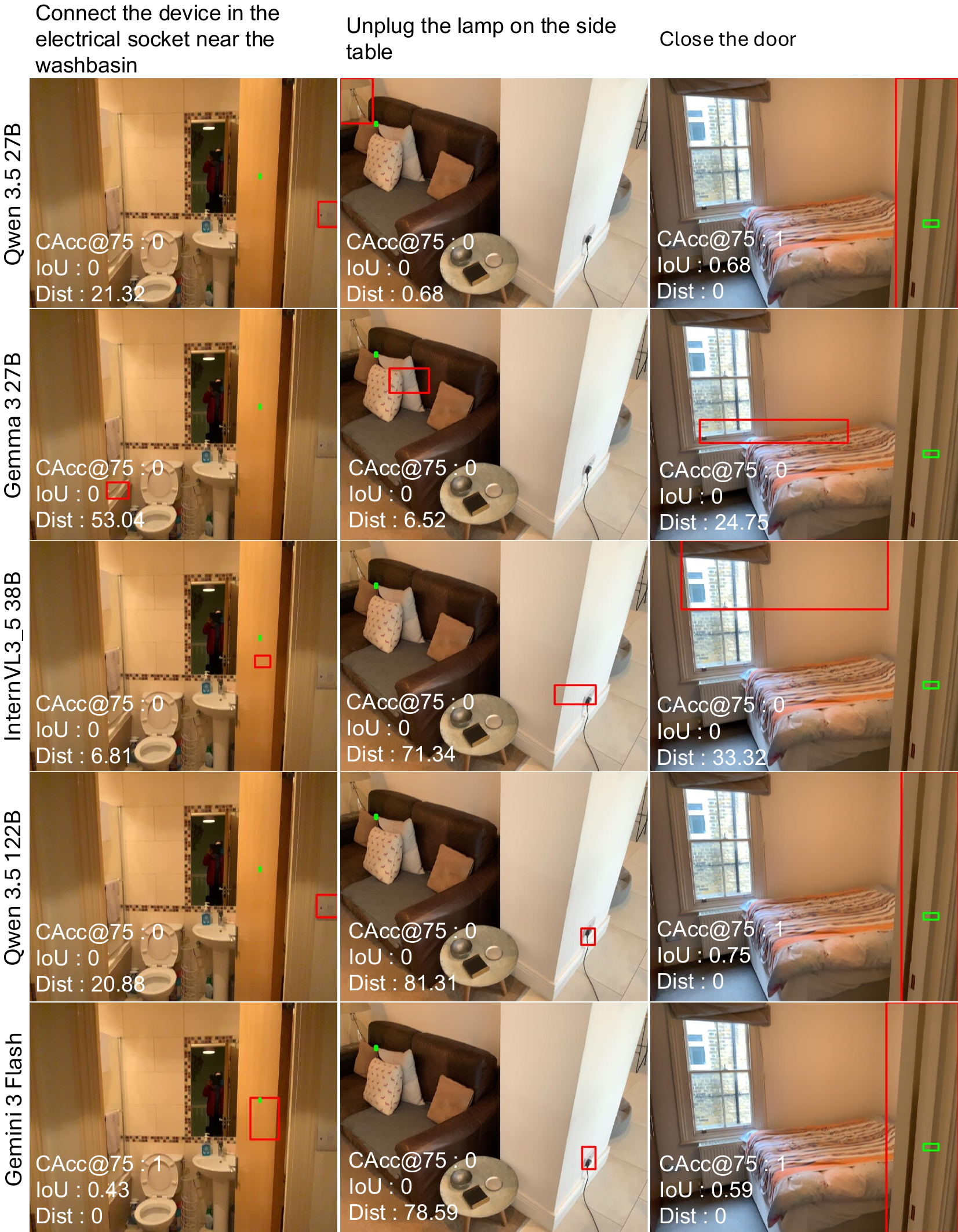}
  \caption{\textbf{Qualitative examples of SceneFunRI.} The red bounding boxes indicate the model predictions, while the green bounding boxes represent the ground-truth locations of the target objects.}
  \label{fig:Qualitative}
\end{figure}

We present qualitative results of multiple language models across different tasks in Figure \ref{fig:Qualitative}. The examples illustrate three distinct task instructions: “Connect the device in the electrical socket near the washbasin,” “Unplug the lamp on the side table,” and “Close the door.” From the visualizations, it is evident that notable differences remain among models in terms of target localization and spatial reasoning capabilities.

We observe that models are easily misled by regions that are irrelevant to the task but exhibit similar visual appearance or semantic associations. For instance, in the first column, both Qwen 3.5 27B and Qwen 3.5 122B incorrectly identify a nearby power switch as the target socket. In the second column, several models again misidentify incorrect sockets as the task target. Only Qwen 3.5 27B successfully localizes the lamp, yet still fails to further identify the correct socket to operate on. These observations suggest that when multiple candidate objects share similar functions or appearances, models may capture partial semantic cues but struggle to correctly integrate object relationships and the fine-grained constraints specified in the task description.

Furthermore, in the third column task “Close the door,” Qwen 3.5 27B, Qwen 3.5 122B, and Gemini 3 Flash are all able to successfully localize the target by highlighting regions near the door frame, indicating a certain level of scene understanding. However, these models still fail to further infer the approximate height and position of the occluded door handle. This suggests that, even when models can identify high-level task objectives, their ability to reason about implicit spatial cues, occluded objects, and fine-grained interaction points remains limited.

\section{Conclusion} 
This paper presents SceneFunRI, a benchmark specifically developed for Reasoning the Invisible. SceneFunRI enables the inference of the locations of invisible functional objects from task instructions. By analyzing model behavior under different prompting strategies, we further reveal the limitations of current VLMs in this setting. Our experimental results indicate that the proposed benchmark captures a particularly challenging aspect of spatial reasoning, highlighting an important capability that remains insufficiently understood and warrants further investigation. Although current models may exhibit latent reasoning abilities, these capabilities are not consistently activated when reasoning about invisible regions. Moreover, the results suggest that current VLMs are more effective at excluding low-probability candidate regions in visible space than at directly inferring the locations of unseen objects. Taken together, these findings reveal that existing VLMs still have limited ability to reason under uncertainty in invisible regions, underscoring a fundamental limitation in their spatial reasoning capabilities.

{
    \small
    \section*{Acknowledgement}
This work was supported in part by National Science and Technology Council, Taiwan, under Grant NSTC 113-2634-F-002-007.
    \bibliographystyle{ieeenat_fullname}
    \bibliography{main}
}


\end{document}